\documentclass[10pt,twocolumn,letterpaper]{article}

\usepackage{wacv}
\usepackage{times}
\usepackage{epsfig}
\usepackage{graphicx}
\usepackage{amsmath}
\usepackage{amssymb}
\usepackage{booktabs}
\usepackage{multirow}
\usepackage{pifont}
\usepackage{hyperref}

\usepackage{array}
\newcolumntype{P}[1]{>{\centering\arraybackslash}p{#1}}
\hypersetup{ colorlinks=True, linkcolor=blue, filecolor=black, urlcolor=blue, }

\def\wacvPaperID{654} % Enter the WACV Paper ID here

\wacvfinalcopy % *** Uncomment this line for the final submission

\begin{document}

%%%%%%%%% TITLE
\title{Unsupervised Video Object Segmentation via Prototype Memory Network}

\author{
	Minhyeok Lee \quad
	Suhwan Cho \quad
	Seunghoon Lee \quad
	Chaewon Park \quad
	Sangyoun Lee$^{*}$ \quad
	%\and 
	\vspace{0.01cm}\\
	Yonsei University \\
	{\tt\small \{hydragon516,chosuhwan,shlee423,chaewon28,syleee\}@yonsei.ac.kr}
	%       \and
	%       $^{2}$Microsoft Research Asia \\
	%       {\tt\small \{wuzhiron,stevelin\}@microsoft.com}
}

\maketitle
\thispagestyle{empty}

%%%%%%%%% ABSTRACT
\begin{abstract}
	Unsupervised video object segmentation aims to segment a target object in the video without a ground truth mask in the initial frame. This challenging task requires extracting features for the most salient common objects within a video sequence. This difficulty can be solved by using motion information such as optical flow, but using only the information between adjacent frames results in poor connectivity between distant frames and poor performance. To solve this problem, we propose a novel prototype memory network architecture. The proposed model effectively extracts the RGB and motion information by extracting superpixel-based component prototypes from the input RGB images and optical flow maps. In addition, the model scores the usefulness of the component prototypes in each frame based on a self-learning algorithm and adaptively stores the most useful prototypes in memory and discards obsolete prototypes. We use the prototypes in the memory bank to predict the next query frame’s mask, which enhances the association between distant frames to help with accurate mask prediction. Our method is evaluated on three datasets, achieving state-of-the-art performance. We prove the effectiveness of the proposed model with various ablation studies.
\end{abstract}

\begin{figure}[t]
	\setlength{\belowcaptionskip}{-24pt}
	\begin{center}
		\includegraphics[width=\linewidth]{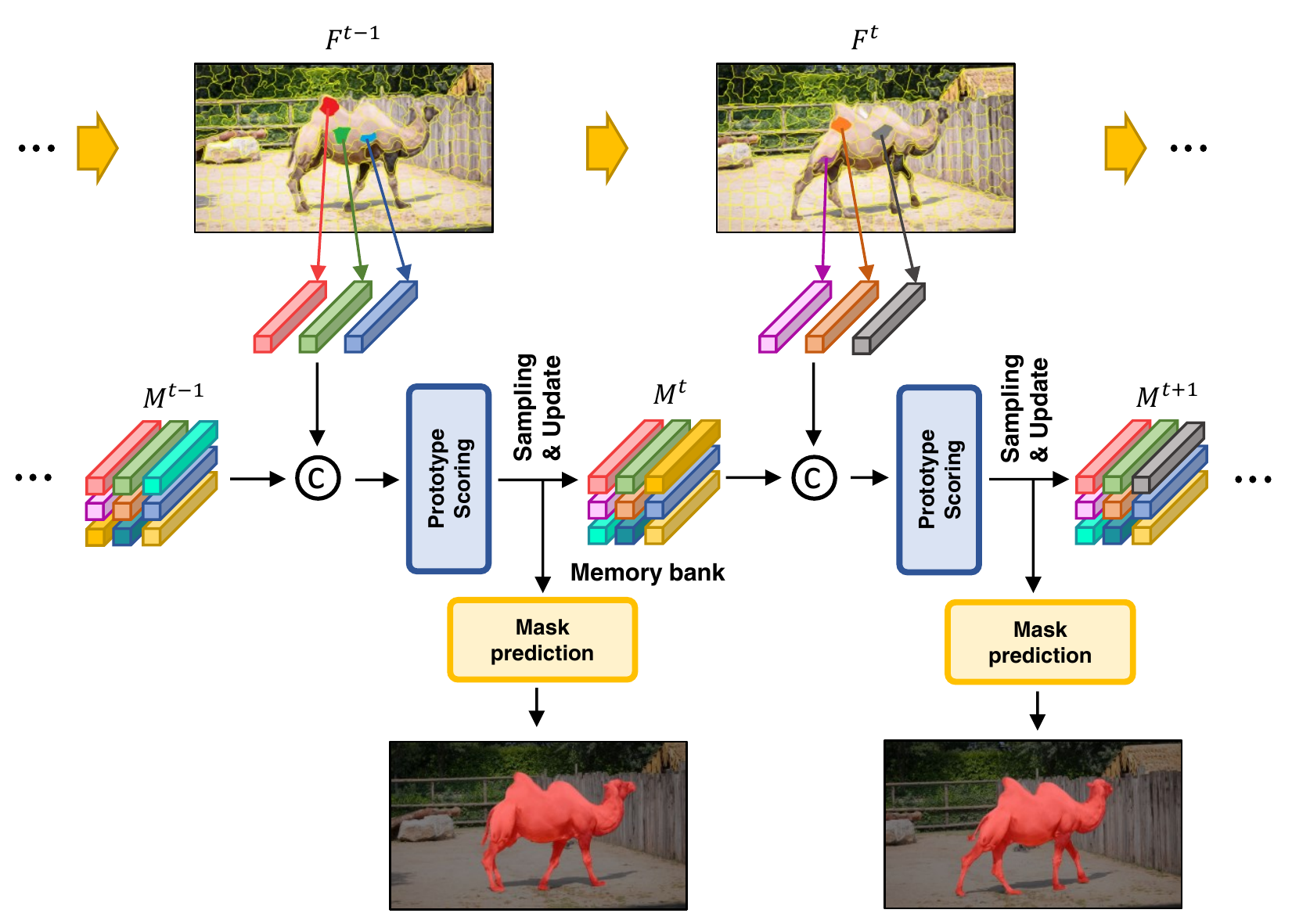}
		\caption{The overall flow of the proposed method. We split the image into superpixels and create prototypes covering each superpixel area. Prototypes are evaluated for their usefulness along with the previous prototypes in the memory bank, and the most useful prototypes are stored in the memory bank for mask prediction in the next sequence.}
		\label{fig:intro}
	\end{center}
\end{figure}

%%%%%%%%% BODY TEXT
\section{Introduction}
Video object segmentation (VOS) aims to delineate pixel-level salient object masks in each frame. VOS is used as preprocessing for video captioning~\cite{wang2018reconstruction}, interactive segmentation~\cite{sofiiuk2020f}, and optical flow estimation~\cite{cheng2017segflow}. It is also widely applied in robotics and autonomous vehicles~\cite{abramov2012depth, liu2020video, maddern20171}.

VOS tasks are divided into semi-supervised~\cite{cho2022pixel, mao2021joint, xie2021efficient} and unsupervised VOS~\cite{fragkiadaki2015learning, tokmakov2017learning, zhou2020motion, ji2021full} depending on whether the ground truth mask of the first frame of the video sequence is provided. To be more specific, semi-supervised VOS aims to track and segment a specified object in the initial frame of a video. However, in an unsupervised VOS task, the model must track and segment the most salient objects without being given a specified mask in the first frame. Therefore, an unsupervised VOS task is very challenging because it is important to search for common objects in the input video sequence and effectively extract their features.

To extract common, consistent features, traditional handcrafted methods~\cite{papazoglou2013fast, faktor2014video, zhou2016video, wang2015saliency, ochs2013segmentation} applied temporal trajectory, saliency prior, and object proposal techniques. However, these methods perform poorly in complex morphological variations of the target object over time and in extreme lighting conditions. To solve this problem, deep-learning-based unsupervised VOS models~\cite{fragkiadaki2015learning, tokmakov2017learning, zhou2020motion, ji2021full, lee2021iteratively, chen2022video} have recently been in the spotlight. In particular, many models~\cite{fragkiadaki2015learning, tokmakov2017learning, zhou2020motion, ji2021full} extract additional motion information from the optical flow and use it as a guideline for common objects. However, because these models generate optical flow maps between two adjacent frames, they ignore the feature associations between long-distance frames, resulting in poor performance. It is also difficult to fuse the two features effectively because of the large domain gap between the RGB image and the optical flow. To solve these problems, Schmidtt~\etal~\cite{schmidt2022d2conv3d} applied 3D convolution to make the model learn long-distance frame dependencies, but this method cannot perform real-time prediction.

We propose a novel prototype memory network (PMN) to address the aforementioned difficulties of unsupervised VOS. Figure~\ref{fig:intro} shows the overall flow of the proposed method. Many studies of segmentation tasks~\cite{tu2018learning, bailoni2022gasp, yang2020superpixel} have shown that preprocessing using superpixels can provide useful features to models and improve performance by clustering image pixels. Therefore, we first divide the RGB images and optical flow maps into superpixels using a simple linear iterative clustering (SLIC) algorithm~\cite{achanta2012slic} to effectively extract various detail and texture information from the RGB images and motion information from the optical flow maps. We then create component prototypes from the superpixel mask, focusing on prototype learning, which is widely used in few-shot segmentation tasks~\cite{dong2018few, wang2019panet, liu2020part}. We also propose a prototype scoring module (PSM) and memory bank to enhance common feature associations between distant frames. The PSM scores the usefulness of the generated prototypes and samples only the most highly scored prototype features. The prototypes selected by the PSM are stored in a memory bank, and these prototypes are combined with the prototypes generated from the next frame image. The PSM updates the memory bank at every frame by giving the prototypes new scores. Therefore, the memory banks can store useful features for target objects from past frames so the model can use them for prediction in future frames. The proposed PSM is trained with self-learning techniques because the usefulness scores for the prototypes are not labeled manually.

We tested our method on three popular datasets: DAVIS16~\cite{perazzi2016benchmark}, FBMS~\cite{ochs2013segmentation}, and YouTube-Objects~\cite{prest2012learning}. These datasets contain various challenging scenarios, and the proposed model achieves state-of-the-art performance on all three datasets. In addition, we performed various ablation studies to prove the effectiveness of the proposed model and show that robust VOS is possible in challenging video sequences.

Our main contributions can be summarized as follows:

\begin{itemize}
	\item We propose a novel PMN to extract detailed information from the RGB image and motion information from the optical flow and strengthen the connectivity between distant frames. Inspired by the prototype learning used in few-shot segmentation tasks, the proposed model generates component prototypes based on the superpixel algorithm.
	
	\item We propose a PSM to score the usefulness of the prototypes generated and update the most highly scored prototypes in memory. The PSM is self-supervised because prototype utility scores cannot be labeled.
	
	\item The proposed network achieves state-of-the-art performance on the DAVIS-16~\cite{perazzi2016benchmark}, FBMS~\cite{ochs2013segmentation}, and YouTube-Objects~\cite{prest2012learning} datasets. Additionally, we demonstrate the effectiveness of the proposed method through various ablation studies.
\end{itemize}

\section{Related Work}

\noindent
\textbf{Unsupervised VOS.} Unsupervised VOS aims to segment eye-catching objects in a video sequence without human intervention. This is an extension of previous single-image salient object detection tasks and is more challenging because it detects common salient objects within a video sequence. Traditional methods~\cite{papazoglou2013fast, faktor2014video, zhou2016video, wang2015saliency, ochs2013segmentation} use motion boundaries, long-term point trajectories, and objectness to segment common objects. However, these methods often fail due to the occlusion of the target object, illumination extremes, and complex foreground and background structures. To perform robust VOS in such challenging situations, deep-learning methods have been in the spotlight recently. In particular, methods~\cite{fragkiadaki2015learning, tokmakov2017learning, zhou2020motion, ji2021full, kim2021aibm} that focus on the motion information of an object perform well in unsupervised VOS tasks. For example, Fragkiadaki~\textit{et al.}~\cite{fragkiadaki2015learning} ranked segment proposals by fusing optical flow and static boundaries. Tokmakov~\textit{et al.}~\cite{tokmakov2017learning} used only optical flow to capture motion cues, but it is difficult to segment static objects with their method due to insufficient detail information. In addition, MATNet~\cite{zhou2020motion} uses motion information to enhance the spatiotemporal object representation. However, this static and motion information fusion method performs poorly with complex moving backgrounds, and there is a problem dependent on the accuracy of the optical flow map.

\begin{figure*}[t]
	\setlength{\belowcaptionskip}{-24pt}
	\begin{center}
		\includegraphics[width=\linewidth]{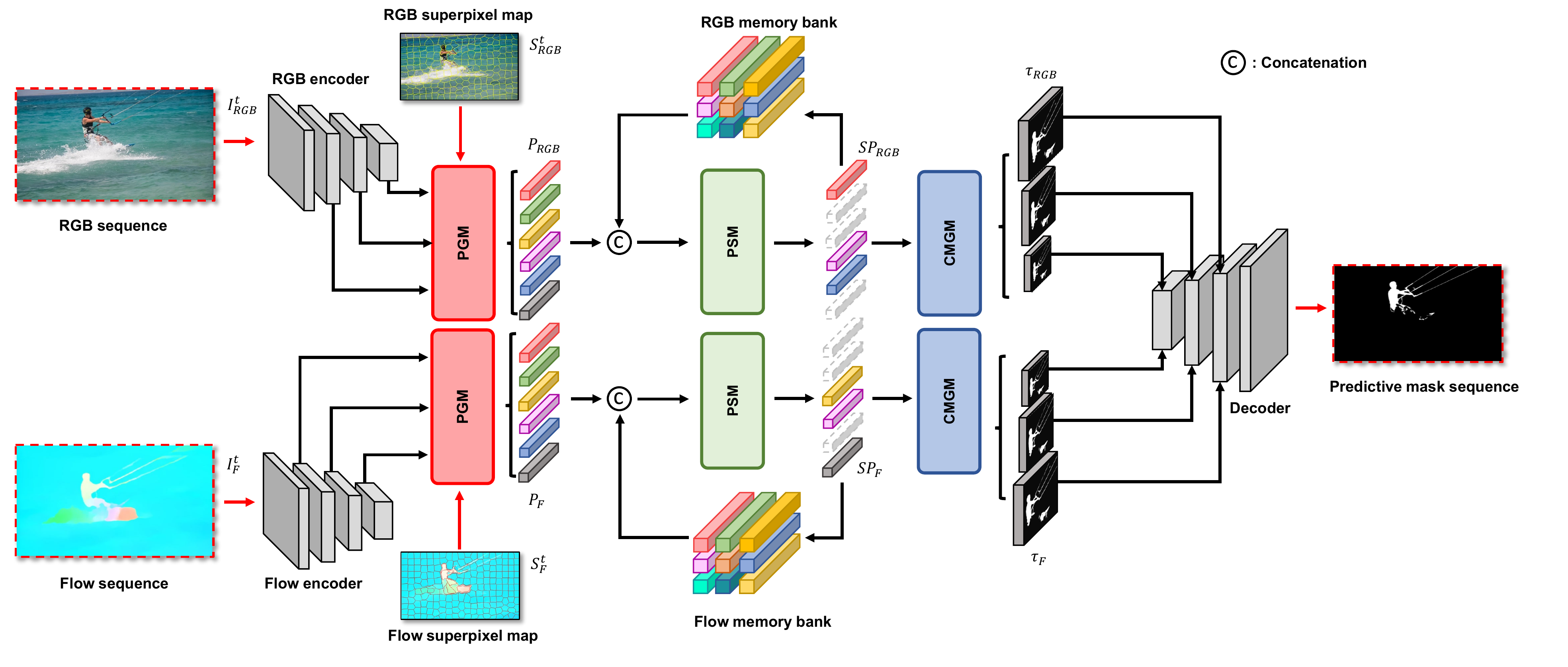}
		\caption{The overall architecture of the proposed prototype memory network (PMN). The prototype generating module (PGM) generates prototypes from images and flow maps. The prototype sampling module (PSM) samples the most useful prototypes by scoring the usefulness of the extracted prototypes. The memory bank stores the sampled prototypes to help predict the mask for the next frame. Finally, the correlation map generating module (CMGM) generates correlation maps from the sampled prototypes.}
		\label{fig:main}
	\end{center}
\end{figure*}

\noindent
\textbf{Prototype Learning.} Prototype learning is a method of learning a metric space in which features can be distinguished by calculating the distance to the prototype representation of each feature. In particular, prototype learning-based deep-learning models~\cite{dong2018few, wang2019panet, liu2020part, yang2020prototype, yu2021location, li2021adaptive} perform well in few-shot segmentation tasks. For example, Wang~\textit{et al.}~\cite{wang2019panet} proposed a model that creates query masks with prototypes generated from support features and then creates support masks with prototypes generated from query features. Yang~\textit{et al.}~\cite{yang2020prototype} proposed a prototype mixture model, effectively fusing foreground and background prototypes. Li~\textit{et al.}~\cite{li2021adaptive} extracted robust features through clustering-based adaptive prototype learning and improved the few-shot segmentation performance.

We propose a memory prototype sampling architecture inspired by prototype learning to extract features of primary objects effectively from video sequences. However, unlike previous few-shot segmentation methods, the proposed method generates various prototypes representing objects by using the SLIC~\cite{achanta2012slic} algorithm. It also improves the unsupervised VOS performance by sampling prototypes of common objects and storing them in a memory bank.

\section{Proposed Approach}
\subsection{Overall Architecture}
Figure.~\ref{fig:main} shows the overall architecture of the proposed PMN. As inputs, the PMN uses an RGB image $\mathbf{I^{t}_{RGB}} \in \mathbb{R} ^ {3 \times H \times W}$ at time $t$ of the video sequence and an optical flow map $\mathbf{I^{t}_{F}} \in \mathbb{R} ^ {3 \times H \times W}$ generated from $\mathbf{I^{t}_{RGB}}$ and $\mathbf{I^{t+1}_{RGB}}$ and their superpixel maps $\mathbf{S^{t}_{RGB}}$, $\mathbf{S^{t}_{F}}$. The proposed model consists of three primary parts: the PGM, PSM, and CMGM. We also create an RGB memory bank and a flow memory bank to store useful prototypes generated at time $t$ and use them for mask prediction at time $t+1$. The PMN also has two encoders for RGB images and flow maps and one decoder.

\begin{figure}[t]
	\setlength{\belowcaptionskip}{-24pt}
	\begin{center}
		\includegraphics[width=\linewidth]{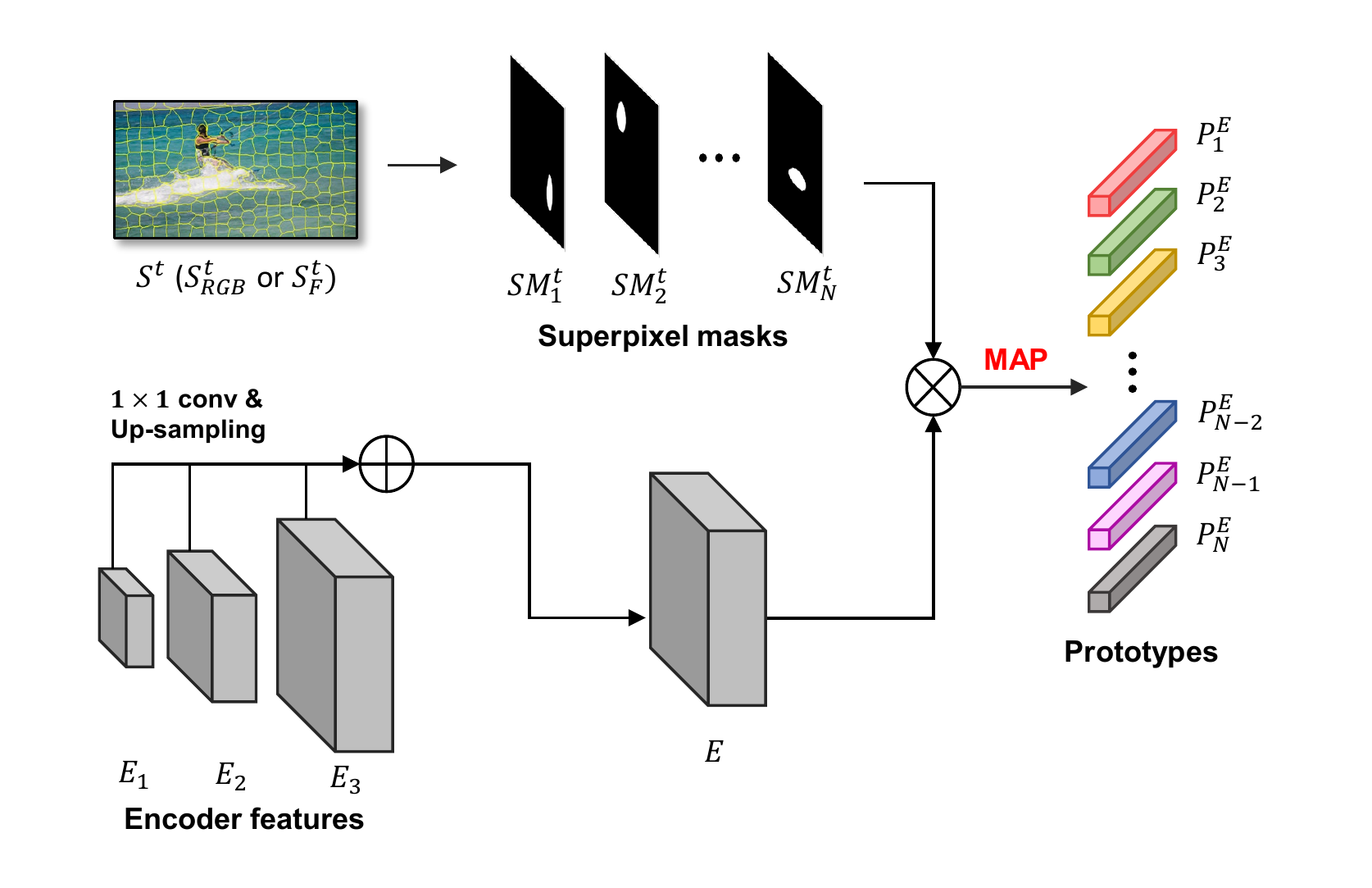}
		\caption{Architecture visualization of the PGM, which creates prototypes that represent the overall features of the subregions that make up the superpixel.}
		\label{fig:PGM}
	\end{center}
\end{figure}

\begin{figure*}[t]
	\setlength{\belowcaptionskip}{-24pt}
	\begin{center}
		\includegraphics[width=\linewidth]{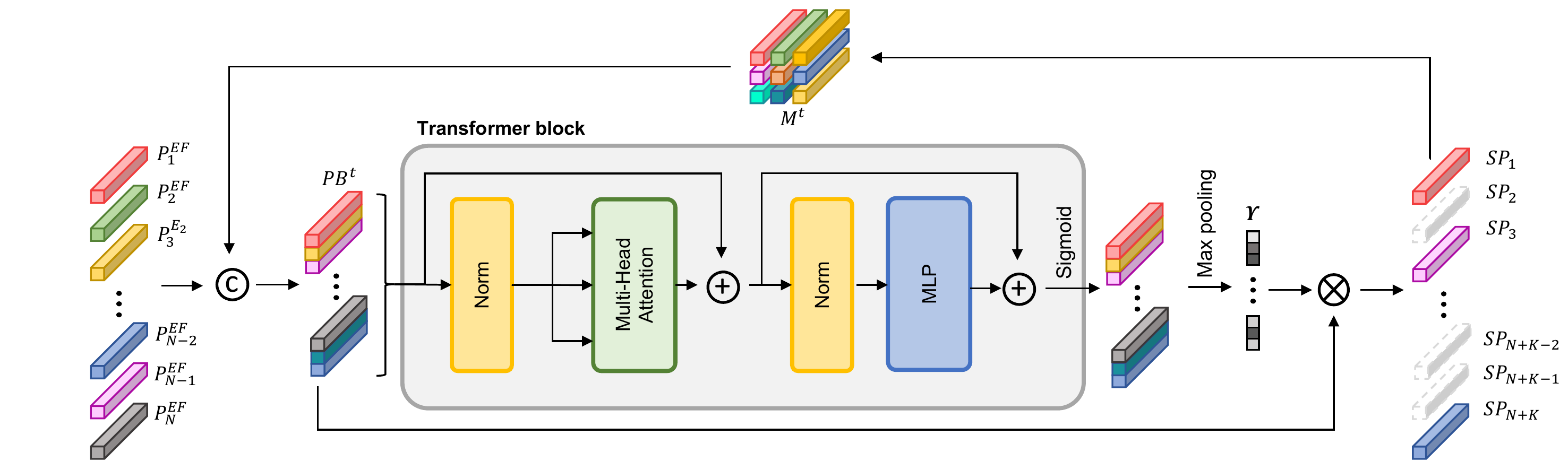}
		\caption{Structure of the proposed PSM and memory bank. The PSM scores the usefulness of the newly extracted prototypes and stores the highest-scoring prototypes in the memory bank.}
		\label{fig:PSM}
	\end{center}
\end{figure*}

\subsection{Prototype Generating Module}
The PGM generates prototypes from the multiscale encoder features $\mathbf{E_{1}}$, $\mathbf{E_{2}}$, and $\mathbf{E_{3}}$. As shown in Figure.~\ref{fig:PGM}, the PGM first integrates $\mathbf{E_{1}}$, $\mathbf{E_{2}}$, and $\mathbf{E_{3}}$ using $1 \times 1$ convolution and upsampling to generate $\mathbf{E} \in \mathbb{R} ^ {C \times \frac{H}{8} \times \frac{W}{8}}$. This architecture, based on a feature pyramid network (FPN)~\cite{lin2017feature}, effectively integrates multiscale features from the encoder. Furthermore, the PGM generates superpixel masks $\mathbf{SM^t_1}$, $\mathbf{SM^t_2}$, ... $\mathbf{SM^t_N} \in \left\{0, 1\right\} ^ {1 \times \frac{H}{8} \times \frac{W}{8}}$ from the superpixel map $\mathbf{S^t}$ at time $t$, where each channel is a binary mask for each superpixel and $N$ is the number of superpixels. To generate prototypes from $\mathbf{E}$, we perform masked average pooling (MAP) using $\mathbf{SM^t}$ as masks, where $\mathbf{SM^t}$ is resized to the same size as $\mathbf{E}$. Consequently, the PGM generates $N$ prototypes $\mathbf{P^{E}_1}$, $\mathbf{P^{E}_2}$, ..., $\mathbf{P^{E}_N} \in \mathbb{R} ^ {1 \times C}$ from $\mathbf{E}$. The PGM process can be summarized as follows: 

\begin{equation}
	\mathbf{P _ { x } ^ { E }} = \frac{ \frac{H}{8} \times \frac{W}{8} } { \sum _ { } ^ { } \mathbf{SM _ { x } ^ { t }} } \times GAP \left ( \mathbf{SM _ { x } ^ { t }} \circ \mathbf{E} \right ),
\end{equation}

\noindent
where $GAP\left(.\right)$ is global average pooling, $\sum _ { } ^ { } \left(.\right)$ is the sum of all pixel values, and $\circ$ is element-wise multiplication. Furthermore, $x = 1$, $2$, ..., $N$. 

Prototype learning in a typical few-shot segmentation task~\cite{dong2018few, wang2019panet, liu2020part, yang2020prototype, yu2021location, li2021adaptive} extracts one representative prototype for each object. However, unlike the previous method, the proposed method extracts various component prototypes based on superpixels so that it can retain various features for RGB images and optical flows.

\subsection{Prototype Scoring Module and the Memory Bank}
Prototypes extracted from the PGM have features that are useful for creating object masks, but they also have features that are not. The PSM samples the most useful prototypes and stores them in a memory bank. In other words, the memory bank stores useful prototypes from the previous $1$ to $t$ time frames, and when a more useful prototype occurs at frame $t+1$, the memory bank is updated. However, because we cannot define the usefulness of the prototypes using the ground truth, the proposed PSM is trained using a self-supervised mechanism. Therefore, the PSM focuses on correlations between prototypes that contain consistent characteristics for salient objects. Figure~\ref{fig:PSM} shows the structure of the proposed PSM and memory bank.

The first step of the PSM generates a prototype block $\mathbf{PB^t}$ by concatenating the prototypes $\mathbf{P^{E}_1}$, $\mathbf{P^{E}_2}$, ..., $\mathbf{P^{E}_N}$ extracted from the encoder and the prototypes $\mathbf{P^{M^t}_1}$, $\mathbf{P^{M^t}_2}$, ..., $\mathbf{P^{M^t}_K}$ in memory bank $M^t$ at time $t$, where $K$ is the number of prototypes in $M^t$. Thus, the size of $\mathbf{PB^t}$ is $\left(N + K \right) \times C$. The next step is a transformer block, inspired by various vision transformer studies~\cite{dosovitskiy2020image, zheng2021rethinking, wang2021end}, to enhance the correlation between prototypes. Unlike general vision transformers, the patch-embedding process is omitted because the input is a prototype block rather than a 2D image. In addition, because the proposed PSM samples useful prototypes from pre-extracted object feature vectors, it is composed of a single transformer layer rather than multiple layers. The transformer block, shown in Figure~\ref{fig:PSM}, consists of a series connection of two layer normalization steps, a multi-head attention layer, a multi-layer perceptron layer, and a sigmoid layer. Therefore, the size of the prototype block with enhanced correlation between prototypes passed through the transformer block is $\left(N + K \right) \times C$, which is equal to the input prototype block size. Finally, the sampling vector $\mathbf{\Upsilon} \in \left[0, 1\right] ^ {\left( N + K \right)}$ to select a useful prototype is calculated by max pooling. Finally, the input prototype block PB is multiplied by this sampling vector to produce a sampled prototype block $\mathbf{SPB^t} \in \mathbb{R} ^ {\left(N + K \right) \times C}$. We also update $M^t$ to $M^{t+1}$ by replacing the prototypes of $M^t$, selecting the top $K$ samples with large $\mathbf{\Upsilon}$ values from among the sampled prototypes $\mathbf{SP_1}$, $\mathbf{SP_2}$, ..., $\mathbf{SP_{N+K}}$. Therefore, the memory bank update process is $\left \{ \mathbf{P _ { n } ^ { M ^ { t+1 } }} \right \} \leftarrow \left \{ \mathbf{SP _ { n }} \right \}$, where $n = 1, 2, ... , K$.

\begin{figure}[t]
	\setlength{\belowcaptionskip}{-24pt}
	\begin{center}
		\includegraphics[width=\linewidth]{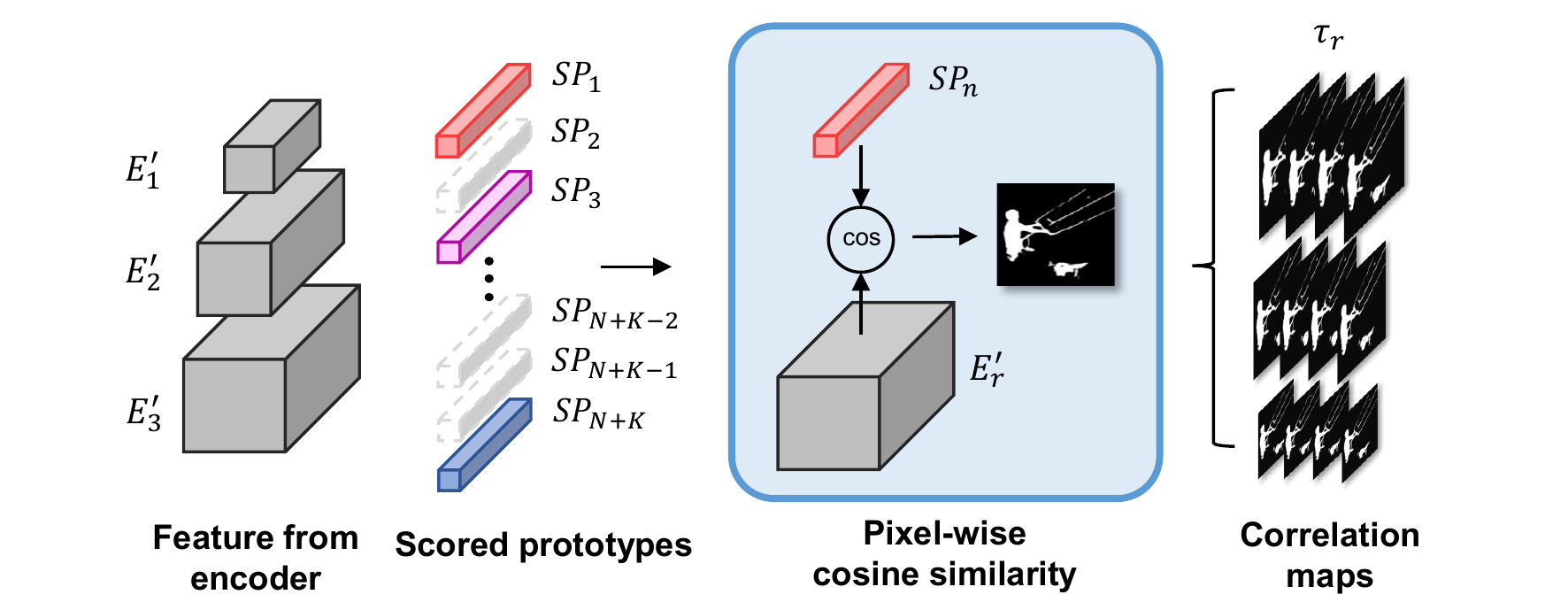}
		\caption{The overall flow of the proposed model. Our model generates and samples component prototypes from the superpixel maps. It also compares the reliability of correlation maps created from component prototypes to generate the predicted mask.}
		\label{fig:CMGM}
	\end{center}
\end{figure}

\subsection{Correlation Map Generating Module}
The CMGM generates correlation maps from the sampled prototypes $\mathbf{SP_1}$, $\mathbf{SP_2}$, ..., $\mathbf{SP_K}$ and encoder features passed through a 1x1 convolutional layer, $\mathbf{E^{'}_{1}}$, $\mathbf{E^{'}_{2}}$, and $\mathbf{E^{'}_{3}}$. Therefore, we compute the pixel-wise cosine similarity between the encoder feature and the sampled prototype, as shown in Figure~\ref{fig:CMGM}. The correlation map $\mathbf{\tau _ { r }}$ generated from $\mathbf{SP_n}$ and $\mathbf{E^{'}_{r}}$ is expressed as follows:

\begin{equation}
	\mathbf{\tau _ { r }}  \left ( x,y \right ) = \textrm{concat}_{n} \left( \frac{ \mathbf{E^{'} _ { r }} \left ( x,y \right ) \cdot \mathbf{SP _ { n }} } { \left \| \mathbf{E^{'} _ { r }} \left ( x,y \right ) \right \| \left \| \mathbf{SP _ { n }} \right \| } \right),
\end{equation}

\noindent
where $\left( x,y \right)$ are the pixel coordinates and $r = 1, 2, 3$ and $\textrm{concat}_{n} \left(.\right)$ is the channel concatenation operator, where $n=1, 2, ..., K$. This process allows the model to generate adaptive correlation maps from critical prototypes.

\subsection{Loss Function}
We optimize the model with object function $L$, where $L$ is the intersection over union (IOU) loss~\cite{lin2019agss} between the predicted saliency map $I_{pred}$ and the ground truth mask $I_{gt}$, expressed as:

\begin{equation}
	L = 1- \frac{ \sum ^ { } \textrm{min} \left ( I _ { pred } \left ( x,y \right ) ,I _ { gt } \left ( x,y \right ) \right ) } { \sum ^ { } \textrm{max} \left ( I _ { pred } \left ( x,y \right ) ,I _ { gt } \left ( x,y \right ) \right ) },
\end{equation}

\noindent
where $\textrm{min} \left (.,. \right )$ and $\textrm{max} \left (.,. \right )$ represent the functions that take two maps as inputs and output the element-wise minimum and maximum, respectively. $\left( x,y \right)$ are the pixel coordinates.

\begin{table*}[t]
	\begin{center}
		\caption{Performance comparison with other state-of-the-art methods on the DAVIS-16~\cite{perazzi2016benchmark} dataset. Higher scores are better. The best and second best are highlighted in \textcolor{red}{red} and \textcolor{blue}{blue}, respectively.
	}
		\label{table:tb1}
		\begin{tabular}{c|c|c|P{1.5cm}|ccc}
			\hline
			Method       & Year                  & Backbone       & CRF                   & $\mathcal{J}$\&$\mathcal{F}$ $\uparrow$                & $\mathcal{J}$-Mean $\uparrow$              & $\mathcal{F}$-Mean $\uparrow$              \\ \hline
			AGS~\cite{wang2019learning}          & CVPR 2019             & ResNet101~\cite{he2016deep}      & \ding{51}                  & 78.6                 & 79.7                 & 77.4                 \\
			COSNet~\cite{lu2019see}       & CVPR 2019             & DeepLabv3~\cite{chen2017rethinking}  & \ding{51}                  & 80.0                 & 80.5                 & 79.4                 \\
			AD-Net~\cite{yang2019anchor}       & ICCV 2019             & ResNet101~\cite{he2016deep}      &                       & 81.1                 & 81.7                 & 80.5                 \\
			AGNN~\cite{wang2019zero}         & ICCV 2019             & DeepLabv3~\cite{chen2017rethinking}      & \ding{51}                  & 79.9                 & 80.7                 & 79.1                 \\
			MATNet~\cite{zhou2020motion}       & AAAI 2020             & ResNet101~\cite{he2016deep}      &                       & 81.6                 & 82.4                 & 80.7                 \\
			WCS-Net~\cite{zhang2020unsupervised}      & ECCV 2020             & EfficientNetv2~\cite{tan2021efficientnetv2} & \ding{51}                  & 81.5                 & 82.2                 & 80.7                 \\
			DFNet~\cite{zhen2020learning}        & ECCV 2020             & DeepLabv3~\cite{chen2017rethinking}      & \ding{51}                  & 82.6                 & 83.4                 & 81.8                 \\
			3DC-Seg~\cite{mahadevan2020making}      & BMVC 2020             & ResNet152~\cite{he2016deep}      &                       & 84.5                 & 84.3                 & 84.7                 \\
			F2Net~\cite{liu2021f2net}        & AAAI 2021             & DeepLabv3~\cite{chen2017rethinking}      &                       & 83.8                 & 83.1                 & 84.4                 \\
			RTNet~\cite{ren2021reciprocal}        & CVPR 2021             & ResNet101~\cite{he2016deep}      & \ding{51}                  & 85.2                 & \textcolor{red}{85.6}                 & 84.7                 \\
			FSNet~\cite{ji2021full}        & ICCV 2021             & ResNet50~\cite{he2016deep}       & \ding{51}                  & 83.3                 & 83.4                 & 83.1                 \\
			TransportNet~\cite{zhang2021deep} & ICCV 2021             & ResNet101~\cite{he2016deep}      &                       & 84.8                 & 84.5                 & 85.0                 \\
			AMC-Net~\cite{yang2021learning}      & ICCV 2021             & ResNet101~\cite{he2016deep}      & \ding{51}                  & 84.6                 & 84.5                 & 84.6                 \\
			CFAM~\cite{chen2022video}         & WACV 2022             & ResNet101~\cite{he2016deep}      &                       & 82.8                 & 83.5                 & 82.0                 \\
			IMP~\cite{lee2021iteratively}          & AAAI 2022             & ResNet50~\cite{he2016deep}       &                       & \textcolor{blue}{85.6}                 & 84.5                 & \textcolor{red}{86.7}                 \\ \hline
			Ours         &  & VGG16~\cite{simonyan2014very}          &  & \textcolor{red}{85.9} & \textcolor{blue}{85.4} & \textcolor{blue}{86.4} \\
			Ours         &                       & VGG16~\cite{simonyan2014very}          & \ding{51}                   &     \textcolor{red}{85.9}          &           \textcolor{red}{85.6}           &         86.2             \\ \hline
		\end{tabular}
	\end{center}
\end{table*}

\section{Experiments}
\subsection{Datasets and Evaluation Metrics}
We perform experiments on the following three popular unsupervised VOS benchmarks to validate the effectiveness of our proposed method: DAVIS-16~\cite{perazzi2016benchmark}, FBMS~\cite{ochs2013segmentation}, and Youtube-Objects~\cite{prest2012learning}. Note that most of the existing unsupervised VOS works use these datasets as the test sets [61, 68, 32]. Thus, we follow the same practice to ensure a fair comparison.

\noindent
\textbf{DAVIS-16.} DAVIS-16~\cite{perazzi2016benchmark} is the most popular unsupervised VOS dataset, consisting of 30 training and 20 validation annotated video sequences. We leverage three evaluation metrics: region similarity $\mathcal{J}$, boundary accuracy $\mathcal{F}$, and overall $\mathcal{J}\&\mathcal{F}$ score, which is the average of the $\mathcal{J}$ and $\mathcal{F}$ scores. $\mathcal{J}$ and $\mathcal{F}$ are defined as follows:

\begin{equation}
	\mathcal{J} = \frac{\sum{\mathbf{S_{pred}} \cap \mathbf{S_{gt}}}}{\sum{\mathbf{S_{pred}} \cup \mathbf{S_{gt}}}},
\end{equation}

\begin{equation}
	\mathcal{F} = \frac{2 \times \textrm{Precision} \times \textrm{Recall}}{\textrm{Precision} + \textrm{Recall}},
\end{equation}

\begin{table}[t]
	\begin{center}
		\caption{Performance comparison with other state-of-the-art methods on the FBMS~\cite{ochs2013segmentation} dataset. Higher scores are better. The best and second best are highlighted in \textcolor{red}{red} and \textcolor{blue}{blue}, respectively.}
		\label{table:tb2}
		\resizebox{\columnwidth}{!}{
			\begin{tabular}{c|c|c|c|c}
				\hline
				Method       & Year      & Backbone  & CRF & $\mathcal{J}$-Mean $\uparrow$ \\ \hline
				SFL~\cite{cheng2017segflow}      & CVPR 2017 & ResNet101~\cite{he2016deep} & \ding{51} & 56.0    \\
				IET~\cite{li2018instance}      & CVPR 2018 & DeepLabv2~\cite{chen2017rethinking} & \ding{51} & 71.9    \\
				PDB~\cite{song2018pyramid}      & ECCV 2018 & ResNet50~\cite{he2016deep}  & \ding{51} & 74.0    \\
				COSNet~\cite{lu2019see}       & CVPR 2019 & DeepLabv3~\cite{chen2017rethinking} & \ding{51} & 75.6    \\
				F2Net~\cite{liu2021f2net}        & AAAI 2021 & DeepLabv3~\cite{chen2017rethinking} &     & 77.5    \\
				AMC-Net~\cite{yang2021learning}      & ICCV 2021 & ResNet101~\cite{he2016deep} & \ding{51} & 76.5    \\
				IMP~\cite{lee2021iteratively}          & AAAI 2022 & ResNet50~\cite{he2016deep}  &     & 77.5    \\ \hline
				Ours         &     &   VGG16~\cite{simonyan2014very}   &     &     \textcolor{blue}{77.7}    \\
				Ours         &     &   VGG16~\cite{simonyan2014very}   & \ding{51} &   \textcolor{red}{77.8}      \\ \hline
			\end{tabular}
		}
	\end{center}
\end{table}

\begin{table}[t]
	\begin{center}
		\caption{Performance comparison with other state-of-the-art methods on the Youtube-Objects~\cite{prest2012learning} dataset. Higher scores are better. The best and second best are highlighted in \textcolor{red}{red} and \textcolor{blue}{blue}, respectively.}
		\label{table:tb3}
		\resizebox{\columnwidth}{!}{
			\begin{tabular}{c|c|c|c|c}
				\hline
				Method  & Year      & Backbone       & CRF & J -Mean \\ \hline
				AGS~\cite{wang2019learning}     & CVPR 2019 & ResNet101~\cite{he2016deep}      & \ding{51} & 69.7    \\
				COSNet~\cite{lu2019see}  & CVPR 2019 & DeepLabv3~\cite{chen2017rethinking}      & \ding{51} & 70.5    \\
				AGNN~\cite{wang2019zero}    & ICCV 2019 & DeepLabv3~\cite{chen2017rethinking}      & \ding{51} & 70.8    \\
				MATNet~\cite{zhou2020motion}  & AAAI 2020 & ResNet101~\cite{he2016deep}      &     & 69.0    \\
				WCS-Net~\cite{zhang2020unsupervised} & ECCV 2020 & EfficientNetv2~\cite{tan2021efficientnetv2} & \ding{51} & 70.9    \\
				RTNet~\cite{ren2021reciprocal}   & CVPR 2021 & ResNet101~\cite{he2016deep}      & \ding{51} & 71.0    \\
				AMC-Net~\cite{yang2021learning} & ICCV 2021 & ResNet101~\cite{he2016deep}      & \ding{51} & \textcolor{blue}{71.1}    \\ \hline
				Ours    &           & VGG16~\cite{simonyan2014very} &     & \textcolor{red}{71.8}    \\
				Ours    &           & VGG16~\cite{simonyan2014very} & \ding{51} & \textcolor{red}{71.8}    \\ \hline
			\end{tabular}
		}
	\end{center}
\end{table}

\begin{figure*}[t]
	\setlength{\belowcaptionskip}{-24pt}
	\begin{center}
		\includegraphics[width=\linewidth]{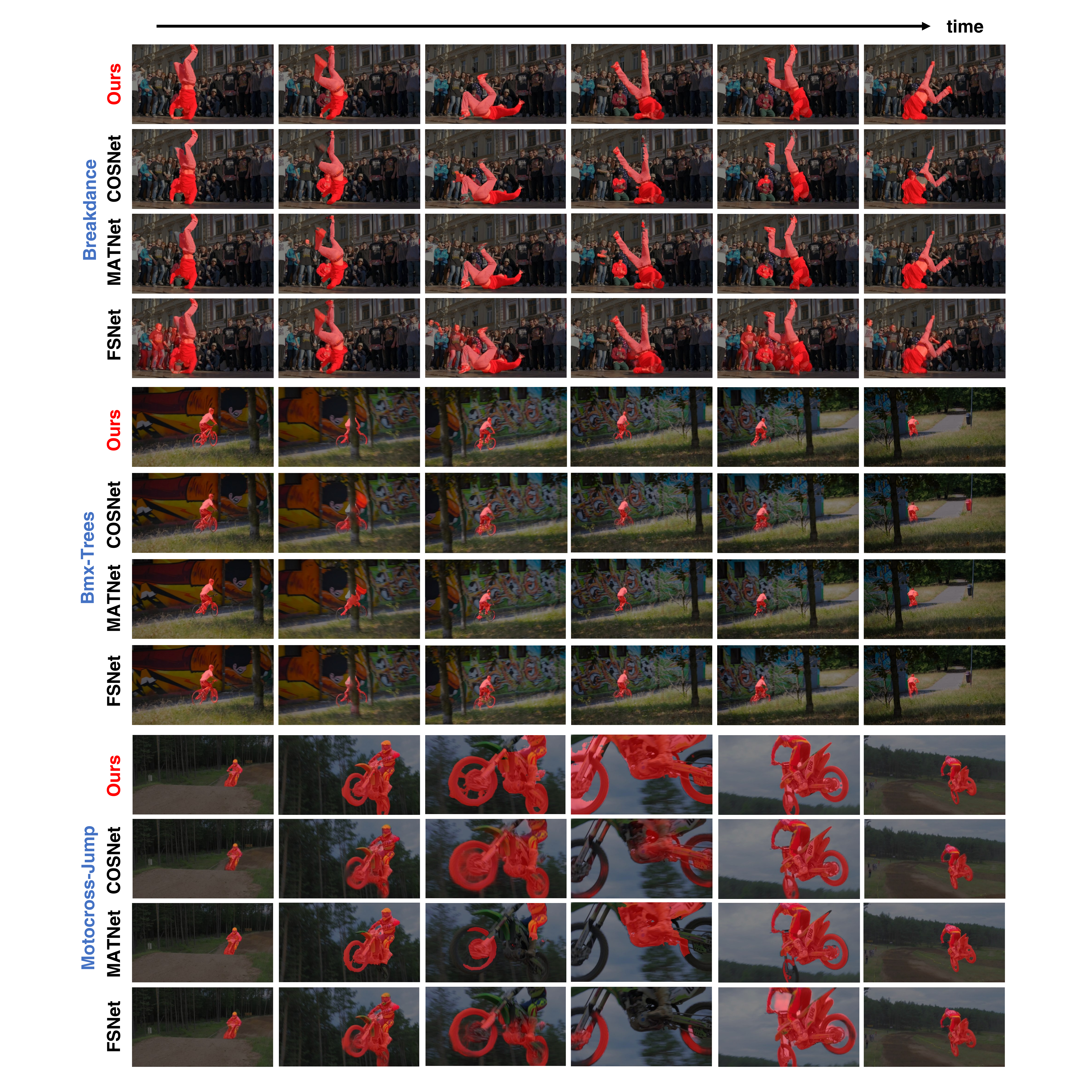}
		\caption{Qualitative comparison of the proposed method with previous state-of-the-art methods, FSNet~\cite{ji2021full}, MATNet~\cite{zhou2020motion}, and COSNet~\cite{lu2019see}. Our method demonstrates robust mask prediction in various challenging videos.}
		\label{fig:result}
	\end{center}
\end{figure*}

\noindent
where $\textrm{Precision} = \sum{\mathbf{S_{pred}} \cap \mathbf{S_{gt}}} / \sum{\mathbf{S_{pred}}}$ and $\textrm{Recall} = \sum{\mathbf{S_{pred}} \cap \mathbf{S_{gt}}} / \sum{\mathbf{S_{gt}}}$.

\noindent
\textbf{FBMS.} FBMS~\cite{ochs2013segmentation} includes 59 video sequences, with 29 are used as the training set and 30 for testing. Following~\cite{yang2019anchor, lu2019see, zhou2020motion}, we used the region similarity $\mathcal{J}$ to evaluate our method on the test set without training.

\noindent
\textbf{YouTube-Objects.} YouTube-Objects~\cite{prest2012learning} contains 126 video sequences of 10 object categories. The ground-truth in YouTube-Objects is sparsely labeled in one of every 10 frames. Following~\cite{yang2019anchor, lu2019see, zhou2020motion}, we use the region similarity $\mathcal{J}$ to evaluate our method on test set without training.

\subsection{Model Training}
We train the model in three steps following the training method of the previous works~\cite{ji2021full, ren2021reciprocal, liu2021f2net, lu2019see}. First, we use a well-known saliency dataset DUTS~\cite{wang2017learning} to pretrain the model to avoid over-fitting. The DUTS~\cite{wang2017learning} dataset consists of a single RGB image and mask. Therefore, only the RGB encoders, PGM, PSM, CMGM, and decoders of the RGB stream in Figure~\ref{fig:main} are pretrained. Second, because the proposed model has perfect symmetry between the RGB stream and the optical flow stream, we apply the pretrained parameters of the RGB stream equally to the optical flow stream. Finally, the entire model is fine-tuned with the DAVIS-16~\cite{perazzi2016benchmark} training set (30 sequences). The optical flow map used for pre-training and fine-tuning is generated using RAFT~\cite{teed2020raft}, a pretrained optical flow estimation model. In addition, 30 memory banks are created, the same number as the sequence of DAVIS-16~\cite{perazzi2016benchmark}, and the prototypes created in each sequence are stored. To prevent the PSM from overfitting by certain prototypes, all memory blocks are reset every epoch.

\subsection{Model Testing}
We follow the standard benchmarks~\cite{fan2019shifting, perazzi2016benchmark} to test our model on the validation set of DAVIS-16~\cite{perazzi2016benchmark}, the test set of FBMS~\cite{ochs2013segmentation}, and the test set of Youtube-Objects~\cite{prest2012learning}. Similar to the training phase, we generate an optical flow map using RAFT~\cite{teed2020raft}, a pretrained optical flow prediction model on three test sets. Furthermore, we initialize the memory banks in each test phase and create empty memory banks for each number of sequences in the test dataset.

\subsection{Implementation Details}
We set the number of superpixels $N$ to 100 and the number of prototypes $K$ in the memory bank to 50. The backbone encoder network is VGG16~\cite{simonyan2014very}, with ImageNet~\cite{deng2009imagenet} pre-trained. All images are uniformly resized to $352 \times 352$ pixels for training and inferring. For network training and fine-tuning, we used the Adam optimizer~\cite{kingma2014adam} with $\beta_1 = 0.9$, $\beta_2 = 0.999$, and $\epsilon = 10^{-8}$. The learning rate decayed from $10^{-4}$ to $10^{-5}$ with the cosine annealing scheduler~\cite{loshchilov2016sgdr}. The total number of epochs was set to 200 with batch size 12. The experiments were conducted on a single NVIDIA RTX 3090 GPU. We implement the proposed method using the open deep-learning framework PyTorch.

\subsection{Results}
In Tables~\ref{table:tb1}, \ref{table:tb2}, and \ref{table:tb3} and Figure~\ref{fig:result}, we compare the performance of the proposed model with previous state-of-the-art methods. Some studies generate a prediction mask and then apply conditional random fields (CRF)~\cite{lafferty2001conditional} to it for post-processing. Therefore, Tables~\ref{table:tb1}, \ref{table:tb2}, and \ref{table:tb3} also present the results of our model applying CRF.

\noindent
\textbf{Quantitative Results.} 
Tables~\ref{table:tb1},~\ref{table:tb2}, and ~\ref{table:tb3} show the quantitative results of the proposed SPSN. The proposed model achieves state-of-the-art performance on all three challenging datasets, even without postprocessing using CRF. We demonstrate the effectiveness of the proposed modules through various ablation studies in the next section.

\noindent
\textbf{Qualitative Results.} Figure~\ref{fig:result} shows visualized results of our model for various challenging video sequences. First, in the breakdance sequence, the proposed model shows robustness in complex background situations with many objects similar in appearance to the target object. It can also be seen in the BMX-Trees sequence that accurate mask generation is possible even when the target object is occluded. Finally, the Motocross-Jump sequence shows that the proposed model is capable of consistent feature extraction, even with extreme scale changes of the objects. These results show that the proposed SPSN extracts the common features of the target object from previous frames and stores them in a memory bank, thereby excluding the influence on non-common objects.

\subsection{Ablation Analysis}
We verified the performance of our model through various ablation studies. Table~\ref{table:tb4} shows the effects of the proposed modules in various combinations.

\begin{table*}[t]
	\begin{center}
		\caption{Performance with different combinations of our contributions on the DAVIS-16~\cite{perazzi2016benchmark} dataset. Higher scores are better. RE and FE represent the encoders for the RGB images and flow maps.}
		\label{table:tb4}
		\begin{tabular}{c|cccccc|ccc}
			\hline
			\multirow{3}{*}{Index} & \multicolumn{6}{c|}{Method}                                                                                                                                                                                                                  & \multirow{3}{*}{$\mathcal{J}$\&$\mathcal{F}$ $\uparrow$} & \multirow{3}{*}{$\mathcal{J}$-Mean $\uparrow$} & \multirow{3}{*}{$\mathcal{F}$-Mean $\uparrow$} \\ \cline{2-7}
			& \multicolumn{2}{c|}{Encoder}                                                 & \multicolumn{2}{c|}{PSMN}                                                    & \multicolumn{1}{c|}{\multirow{2}{*}{Memory Bank}} & \multirow{2}{*}{CMGM}      &                       &                         &                         \\ \cline{2-5}
			& RE\&PGM                    & \multicolumn{1}{c|}{FE\&PGM}                    & Transformer                & \multicolumn{1}{c|}{MLP}                        & \multicolumn{1}{c|}{}                             &                            &                       &                         &                         \\ \hline      
			(a)                    & \ding{51} & \multicolumn{1}{c|}{}                           &                            & \multicolumn{1}{c|}{}                           & \multicolumn{1}{c|}{}                           &                            &  82.5 & 82.3 & 83.7 \\
			(b)                    & \ding{51} & \multicolumn{1}{c|}{\ding{51}} &                            & \multicolumn{1}{c|}{}                           & \multicolumn{1}{c|}{}                           &                            & 83.3 & 83.0 & 83.6 \\
			(c)                    & \ding{51} & \multicolumn{1}{c|}{\ding{51}} &                            & \multicolumn{1}{c|}{\ding{51}} & \multicolumn{1}{c|}{}                           & \ding{51} & 84.1 & 84.0 & 83.9 \\
			(d)                    & \ding{51} & \multicolumn{1}{c|}{\ding{51}} &                            & \multicolumn{1}{c|}{\ding{51}} & \multicolumn{1}{c|}{\ding{51}} & \ding{51} & 85.0 & 84.9 & 85.1 \\
			(e)                    & \ding{51} & \multicolumn{1}{c|}{\ding{51}} & \ding{51} & \multicolumn{1}{c|}{}                           & \multicolumn{1}{c|}{}                           & \ding{51} & 84.5 & 84.4 & 84.6 \\
			(f)                    & \ding{51} & \multicolumn{1}{c|}{\ding{51}} & \ding{51} & \multicolumn{1}{c|}{}                           & \multicolumn{1}{c|}{\ding{51}} & \ding{51} & \textbf{85.9} & \textbf{85.4} & \textbf{86.4} \\ \hline
		\end{tabular}
	\end{center}
\end{table*}

\begin{table}[h]
	\begin{center}
		\caption{Statistical comparisons of the prototype sampling methods on the DAVIS-16~\cite{perazzi2016benchmark} dataset. Higher scores are better.}
		\label{table:tb5}
		\resizebox{1\columnwidth}{!}{
			\begin{tabular}{c|c|ccc}
				\hline
				Index & Method     & $\mathcal{J}$\&$\mathcal{F}$ $\uparrow$ & $\mathcal{J}$-Mean $\uparrow$ & $\mathcal{F}$-Mean $\uparrow$ \\ \hline
				(a)   & Random     & 85.0 & 85.0 & 85.1 \\
				(b)   & Grid       & 85.2 & 85.1  & 85.3 \\
				(c)   & Superpixel & \textbf{85.9} & \textbf{85.4} & \textbf{86.4} \\ \hline
			\end{tabular}
		}
	\end{center}
\end{table}

\noindent
\textbf{PSMN and the Memory Bank.} Indices (c), (d), (e), and (f) in Table~\ref{table:tb4} show the effects of the proposed PSM and memory banks. In addition, we compared the performance of the proposed transformer block with the simple multi layer perceptron (MLP). Table~\ref{table:tb4} shows that using the transformer block proposed by PSMN is more effective than using a simple MLP. These results show that the self-attention mechanism of the transformer block in PSMN can enhance the correlation between prototypes and extract useful features effectively. Furthermore, it shows a significant performance improvement when using the proposed memory bank. This is because the PSM and memory banks can recursively sample and store useful prototypes for accurate mask prediction.

\noindent
\textbf{Effects of the Superpixel Algorithm.} Table~\ref{table:tb5} shows the performance of different prototype extraction methods on the DAVIS-16~\cite{perazzi2016benchmark} dataset. The random sampling method (a) considers random pixels in an image as superpoints and generates a prototype from that coordinate. The grid method (b) generates prototypes using evenly divided square masks from an image. We set the number of prototypes to the same value, $N = 100$. As shown in Table~\ref{table:tb5}, our proposed superpixel-based component sampling method outperformed the other methods, demonstrating its strong ability to capture the common features of the video sequence.

\begin{figure}[t]
	\setlength{\belowcaptionskip}{-24pt}
	\begin{center}
		\includegraphics[width=\linewidth]{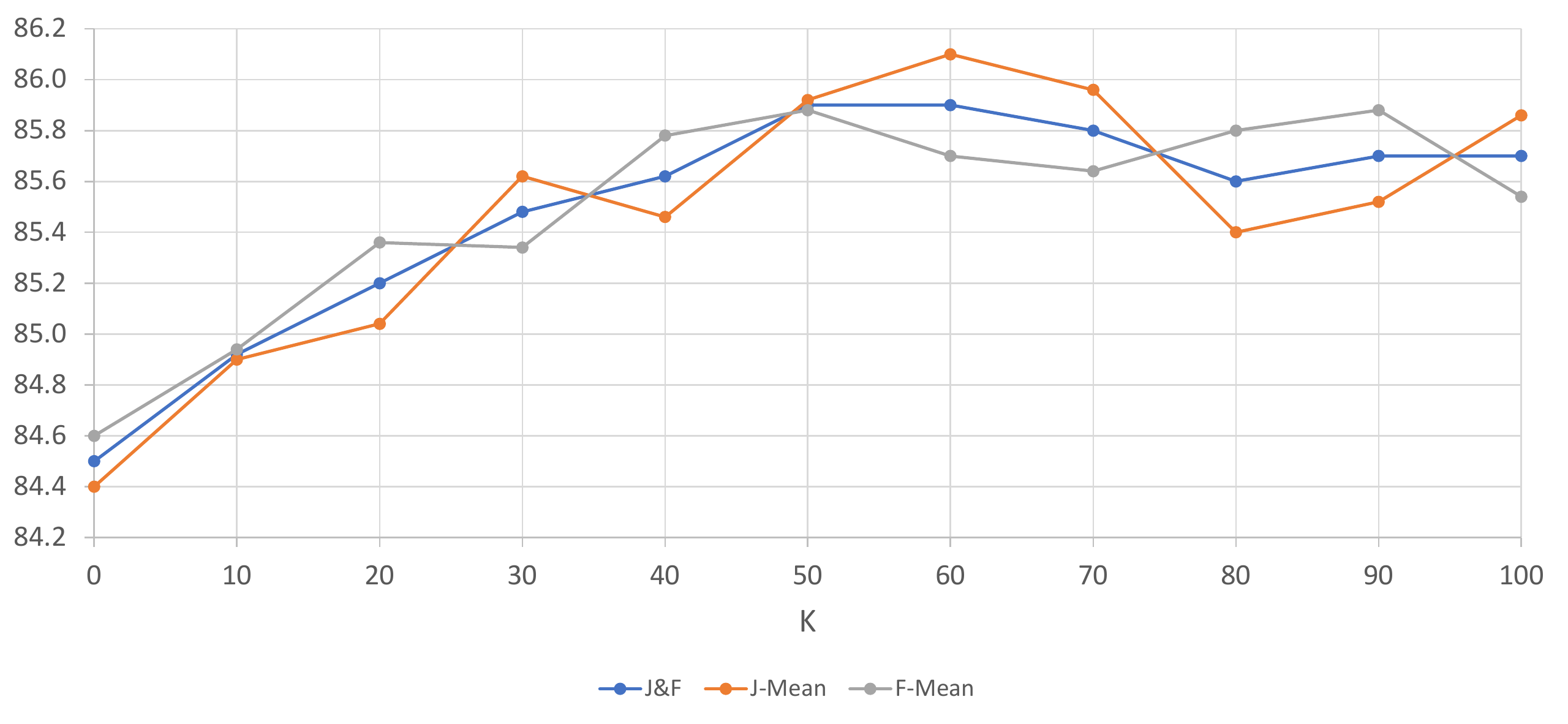}
		\caption{Comparison of performance characteristics with respect to $K$ on the DAVIS-16~\cite{perazzi2016benchmark} dataset. Setting $K = 0$ is the same as not using the memory bank.}
		\label{fig:graph}
	\end{center}
\end{figure}

\begin{figure}[t]
	\setlength{\belowcaptionskip}{-24pt}
	\begin{center}
		\includegraphics[width=\linewidth]{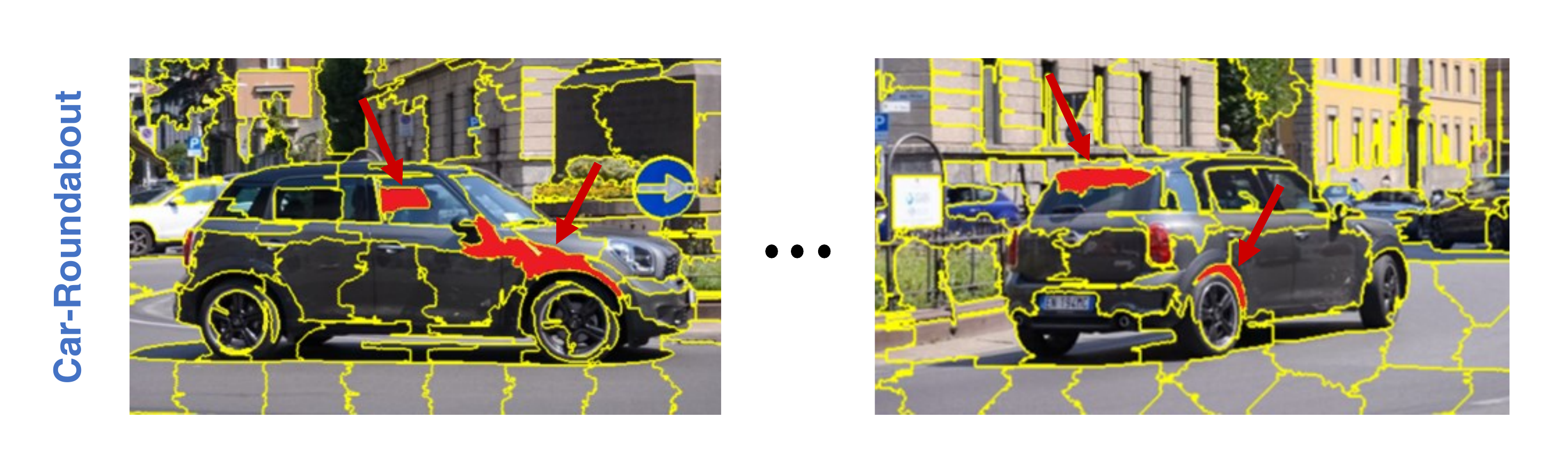}
		\caption{Visualization of two prototypes with the highest score added to the memory bank during testing in Car-Roundabout class. Superpixel areas marked in red indicate new additions to the memory bank in each sequence.}
		\label{fig:last}
	\end{center}
\end{figure}

\noindent
\textbf{Size of the memory bank} Figure~\ref{fig:graph} compares the $\mathcal{J}$\&$\mathcal{F}$ $\uparrow$ scores on DAVIS-16~\cite{perazzi2016benchmark} according to the maximum number of prototypes $K$ stored in the memory bank. The results show the model performed best at $K=50$, with little change in performance when $K$ was greater than 50. This shows that if the capacity of the memory bank is above a certain level, it is possible to hold a sufficient number of prototypes for target object extraction. Furthermore, we visualize the location of newly added prototypes in Figure~\ref{fig:last}. As shown in Figure~\ref{fig:last}, newly added prototypes are generally located on the salient object, showing that the proposed model can sample useful prototypes.

\section{Conclusion}
In this paper, we propose a novel PMN architecture for unsupervised VOS tasks. The PMN extracts appearance features from RGB images and motion features from a flow map by generating component prototypes from the inputs. Furthermore, the model scores the usefulness of the prototypes in each frame and adaptively stores the most conducive prototypes in memory and removes obsolete prototypes. Prototypes stored in the memory bank enhance the mask prediction performance by emphasizing the connectivity between distant frames. Our model was evaluated on three popular datasets, achieving state-of-the-art performance.

{\small
\bibliographystyle{ieee_fullname}
\bibliography{egbib}
}

\end{document}